# Prediction of Flow Characteristics in the Bubble Column Reactor by the Artificial Pheromone-Based Communication of Biological Ants

Shahab Shamshirband [1,2], Meisam Babanezhad [3,] Amir Mosavi [4,5], Narjes Nabipour [6], Eva Hajnal [7], Laszlo Nadai [5], Kwok-Wing Chau [8]

[1] Department for Management of Science and Technology Development, Ton Duc Thang University, Ho Chi Minh City, Vietnam

[2] Faculty of Information Technology, Ton Duc Thang University, Ho Chi Minh City, Vietnam

[3] Department of Energy, Faculty of Mechanical Engineering, South Tehran Branch, Islamic Azad University, Tehran, Iran

[4] School of the Built Environment, Oxford Brookes University, Oxford OX3 0BP, UK; a.mosavi@brookes.ac.uk

[5] Kalman Kando Faculty of Electrical Engineering, Obuda University, Budapest-1034, Hungary: amir.mosavi@kvk.uni-obuda.hu

[6] Institute of Research and Development, Duy Tan University, Da Nang 550000, Vietnam

[7] Alba Regia Technical Faculty, Obuda University, Szekesfehervar, Hungary

[8] Department of Civil and Environmental Engineering, Hong Kong Polytechnic University, Hung Hom, Hong Kong, China

## Abstract

In order to perceive the behavior presented by the multiphase chemical reactors, the ant colony optimization algorithm was combined with computational fluid dynamics (CFD) data. This intelligent algorithm creates a probabilistic technique for computing flow and it can predict various levels of three-dimensional bubble column reactor (BCR). This artificial ant algorithm is mimicking real ant behavior. This method can anticipate the flow characteristics in the reactor using almost 30 % of the whole data in the domain. Following discovering the suitable parameters, the method is used for predicting the points not being simulated with CFD, which represent mesh refinement of Ant colony method. In addition, it is possible to anticipate the bubble-column reactors in the absence of numerical results or training of exact values of evaluated data. The major benefits include reduced computational costs and time savings. The results show a great agreement between ant colony prediction and CFD outputs in different sections of the BCR. The combination of ant colony system and neural network framework can provide the smart structure to estimate biological and nature physics base phenomena. The ant colony optimization algorithm (ACO) framework based on ant behavior can solve all local mathematical answers throughout 3D bubble column reactor. The integration of all local answers can provide the overall solution in the reactor for different characteristics. This new overview of modelling can illustrate new sight into biological behavior in nature.

# 1-Introduction

Multiphase bubble column reactor (BCR) types are highly important for different industries because of their applications and efficiency (Kumar, Degaleesan, Laddha, & Hoelscher, 1976; H Li and Prakash, 2002; Schäfer, Merten, & Eigenberger, 2002). A BCR's structure is composed of a cylindrical vessel with a gas distributor at the bottom section so that the gas bubbles are fed into the reactor (Bouaifi, Hebrard, Bastoul, & Roustan, 2001; Dhotre, Ekambara, & Joshi, 2004; Lefebvre and Guy, 1999; Shah, Kelkar, Godbole, & Deckwer, 1982). Therefore, the gas is sparked in other phases for separation or chemical reaction. Moreover, this phase may have two forms; i.e., liquid-solid mix and liquid phase (Cho, Woo, Kang, & Kim, 2002; Kantarci, Borak, & Ulgen, 2005; Pino et al., 1992; M. Pourtousi, Sahu, & Ganesan, 2014). The BCR is particularly beneficial in petrochemical, chemical, metallurgical, and biochemical industries, and they are utilized as multiple reactors and contactors since these fluid-structure domains give a large surface area (Bombač, Rek, & Levec, 2019; Rieth and Grünewald, 2019; Shi et al., 2019; Shu, Vidal, Bertrand, & Chaouki, 2019). The BCRs in different industries such as pharmaceutical or biochemical are used in the processes that involve reactions such as chlorination, oxidation, polymerization, hydrogenation, and alkylation, which are advantageous for the production of synthetic fuels(Chen, Hasegawa, Tsutsumi, Otawara, & Shigaki, 2003; Ruzicka, Zahradnık, Drahoš, & Thomas, 2001; Sokolichin and Eigenberger, 1994; S. Wang et al., 2003). The Fischer-Tropsch process is considered as a major application of the mentioned reactors in the chemical industries (Prakash, Margaritis, Li, & Bergougnou, 2001). It is the process of indirect coal liquefaction, resulting in various kinds of fuels like synthetic fuels, methanol synthesis, and transportation fuels(Chuntian and Chau, 2002; Maalej, Benadda, & Otterbein, 2003; Rabha, Schubert, & Hampel, 2013). The production of these kinds of fuels is environmentally advantageous compared to the fuels derived from petroleum (Behkish, Men, Inga, & Morsi, 2002; Kantarci, et al., 2005; Michele and Hempel, 2002). The BCRs are extensively used due to their specific operation and design. The high heat transfer coefficients are characteristics of the bubble columns (Buwa and Ranade, 2003; Kantarci, et al., 2005; Krishna and Van Baten, 2003; Leonard, Ferrasse, Boutin, Lefevre, & Viand, 2015; Luo, Lee, Lau, Yang, & Fan, 1999). As the advantage of the bubble columns, it can be stated that a catalyst or other packing chemical components are able to stay a long period even though they are extensively used(Asil, Pour, & Mirzaei; Kannan, Naren, Buwa, & Dutta, 2019; Liu and Luo, 2019; Shi, Yang, Li, Zong, & Yang, 2019; Xin, Zhang, He, & Wang). Also, it is possible to add or remove the online catalyst easily(Deen, Solberg, & Hjertager, 2000; Díaz et al., 2008; Masood and Delgado, 2014; Shimizu, Takada, Minekawa, & Kawase, 2000; Thorat and Joshi, 2004). Thus, the bubble columns are used in biochemical and chemical industries. In order to get effective BCRs, it is necessary to consider their design scale(Krishna, Baten, & Urseanu, 2001; Masood, Khalid, & Delgado, 2015). Hence, if the reactors are improved by computation and simulation of the column's hydrodynamics, then a perfect understanding concerning the process can be provided (M Pourtousi, Ganesan, & Sahu, 2015; Razzaghian, Pourtousi, & Darus, 2012; Verma and Rai, 2003). Various numerical methods are available for estimation of the multiphase flow in the BCRs. Nevertheless, the scholars have difficulties in the simulation of the full gas movement (Besagni, Guédon, & Inzoli, 2018; Hanning Li and Prakash, 2001; Silva, d'Ávila, & Mori, 2012). In order to numerically simulate complex turbulence behavior in the two-phase reactor, often the supercomputers provide the opportunity to calculate the liquid flow in the very complicated geometries. In experimental observation, If the fluid flow is needed to be measured during operation, because of the requirement for the high-speed microscopic cameras and modern probes, it is not economical (Besagni, Guédon, & Inzoli, 2016; Clift, 1978; M. Pourtousi, Zeinali, Ganesan, & Sahu, 2015; Rzehak and Krepper, 2013; H. Wang et al., 2014). Moreover, the other constraint of the approach in the prediction of large BCRs is related to the computational costs at varying operational conditions and different times (Cartland Glover, Blažej, Generalis, & Markoš, 2003; Ekambara, Dhotre, & Joshi, 2005; Hecht and Grünewald, 2019; Jamialahmadi and Müller-Steinhagen, 1992; Joshi, 2001). These limitations gave way to the application of the intelligent algorithms for simulation of BCRs (Burns, Frank, Hamill, & Shi, 2004; Buwa, Deo, & Ranade, 2006; Mohammad Pourtousi, 2016; Xing, Wang, & Wang, 2013).

Support vector machines (Moazenzadeh, Mohammadi, Shamshirband, & Chau, 2018), neural networks (Taherei Ghazvinei et al., 2018), simulated annealing, and evolutionary algorithms are some of the soft computing approaches that can be applied for predicting and simulating the chemical processes(Mahmoud and Ben-Nakhi, 2007; Ozsunar, Arcaklıoglu, & Dur, 2009; Sudhakar, Balaji, & Venkateshan, 2009). This system can direct the complicated relationships(Saleem, Di Caro, & Farooq, 2011). Using this approach, a smart way is provided for the estimation of the complicated mechanisms in engineering. A suitable example in this regard is the regulation of robotic movements in risky cases(Burns, et al., 2004; Krishna, Urseanu, Van Baten, & Ellenberger, 1999; Rampure, Kulkarni, & Ranade, 2007). Thus, this approach is useful in order to control the robots in the cases that the chemical reactions may be dangerous for the people(Buwa, et al., 2006; Simonnet, Gentric, Olmos, & Midoux, 2007; Xing, et al., 2013).

As mentioned, soft modeling approaches pursue a smart process; thus, it is useful in decision-making because of its comprehensiveness and complex algorithm(Berrichi, Yalaoui, Amodeo, &

Mezghiche, 2010). In addition, they can be devoid of various errors including the accuracy in monotonous conditions. In addition, using the different inputs and output procedure is beneficial when the output-input association is inherently meaningful(Lu and Liu, 2013). Therefore, the method's learning process is completely dependent on the data both for experimental or simulated cases (Babanezhad, Rezakazemi, Hajilary, & Shirazian; A. Mosavi, S. Shamshirband, E. Salwana, K.-w. Chau, & J. H. Tah, 2019). The recent research works have been mainly focused on a specific dimension of soft computing methods used for flow patterns production in the BCRs. According to the research works, the relationship between the machine learning and CFD results in important concepts for the computation of different properties of BCRs. A number of researchers, e.g., Mohammad Pourtousi (2016) used different type of big data in the bubble column reactor in the machine learning algorithm and they predicted pattern recognition of gas and liquid flow in the BCR (Fotovatikhah et al., 2018; Yaseen, Sulaiman, Deo, & Chau, 2018). In this study, ant colony method is combined to predict the flow pattern in the BCR. The application of ant colony algorithm is an appropriate alternative rather than using the CFD approach, which is costly in terms of computation, for the flow simulation in BCRs. In this study, the flow characteristics were trained in the BCR by pheromone-based communication of biological ants and compare the results with existing CFD data(Marco Dorigo and Gambardella, 1997; Xu, Chen, Zhu, & Wang, 2010). As a combination of optimization methods and fuzzy system have not been fully used to simulate biological and physics-based phenomena. In this paper, ant optimization method with fuzzy system was used to predict continuous data. For the first time, the optimization method is used as a solver of machine learning to simulate bubble column characteristics.

## 2-Method

For the simulation of bubbling flow in the BCR, the Eulerian–Eulerian approach is used throughout the domain. This method can simulate the fraction of each phase in the domain and it is based on ensemble-averaged mass and momentum transport equations for each phase. For solving fluid flow in the BCR firstly the continuity equation was computed as follows:

$$\frac{\partial}{\partial t}(\rho_k \epsilon_k) + \nabla(\rho_k \epsilon_k u_k) = 0 \tag{1}$$

Momentum transfer equation:

$$\frac{\partial}{\partial t}(\rho_k \epsilon_k u_k) + \nabla(\rho_k \epsilon_k u_k u_k) = -\nabla(\epsilon_k \tau_k) - \epsilon_k \nabla p + \epsilon_k \rho_k g + M_{I,k} \tag{2}$$

The total interfacial force scheme between the main phases are mainly drag and turbulent dispersion force. The overall forcing scheme is written as:

$$M_{I,L} = -M_{I,G} = M_{D,L} + M_{TD,L} \tag{3}$$

The details description of interfacial force methods that are utilized used in this investigation can be observed in (A. Mosavi, S. Shamshirband, E. Salwana, K. W. Chau, & J. H. M. Tah, 2019; Tabib, Roy, & Joshi, 2008). For calculation of the turbulence flow characteristics the k–ε model is utilized for calculation of turbulence behavior in the bubble column reactor. All turbulence model parameters are similar to k-e model (Mohammad Pourtousi, 2016).

### *2.1 Geometrical structure*

A BCR with a height of 2.6 m and diameter of 0.288 m, is used, and a single sparger point is used at the bottom of the column with 0.5 m height. (see Fig.1) The details description of boundary conditions such as slip boundary conditions and degassing pressure at the surface of the column in this investigation can be observed in Pfleger and Becker (2001). The source point boundary condition used for a single Sparger is identical to Tabib, et al. (2008) and A. Mosavi, et al. (2019).

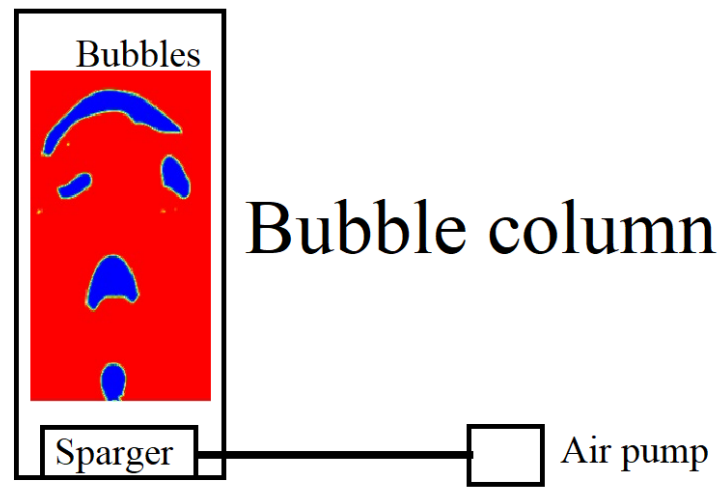

Figure1: Schematic of the bubble column reactor and sparging bubble through the Sparger.

## 2.2 -Grid

A structured hexahedral grid is utilized for calculation of the whole fluid-structure and the interaction between liquid and gas.

## 2.3 -ANT Colony

In this study, the ant colony optimization method (ACO) is a technique for solving big data with complicated problem structures that can be decreased to discover good paths through graphs.

Intelligent Ants or artificial algorithms of ant method stand for multi-agent methods to mimic the real behavior of ants. In this study, this method was used to predict the gas-liquid flow pattern in the column. More description about this method can be found in (Baker and Ayechew, 2003; Bell and McMullen, 2004; Blum, 2005; Castillo, Neyoy, Soria, García, & Valdez, 2013; M Dorigo, Birattari, & Stützle; Marco Dorigo and Blum, 2005; T. Li, Sun, Sattar, & Corchado, 2014; Maroosi and Amiri, 2010; McMullen, 2001; Mocholi, Jaen, Catala, & Navarro, 2010; Mohan and Baskaran, 2012; Mullen, Monekosso, Barman, & Remagnino, 2009; Rao, Srinivasan, & Venkateswarlu, 2010; Suganthi and Samuel, 2012; Tian, Ma, & Yu, 2011; Valdez, Melin, & Castillo, 2014; Yu, Yang, & Yao, 2009). (see Fig.2)

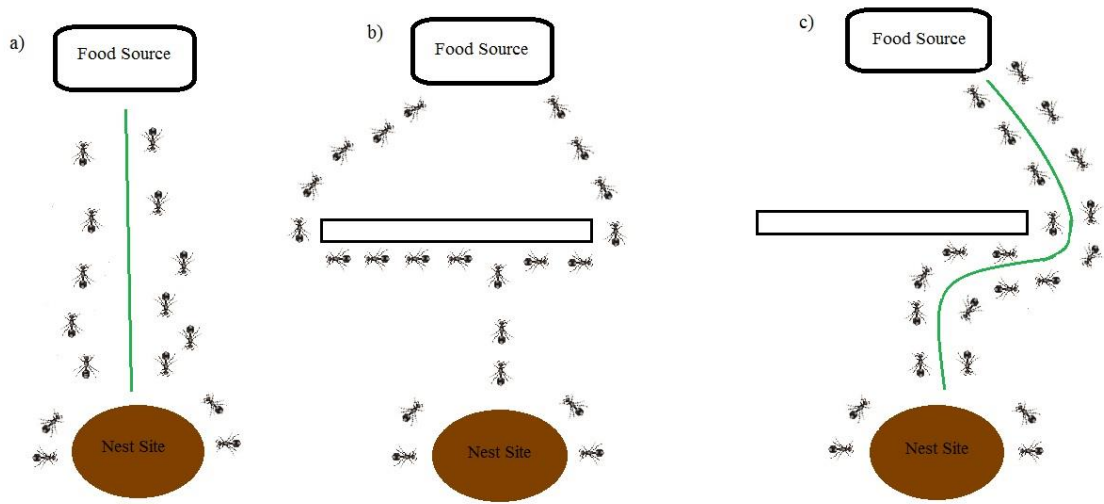

Figure 2: a) Ants food-finding schematic; b) Ants with an obstacle (starting problem); b) Ants with an obstacle (best solution).

## 3-results

In the present study, through simulating a cylindrical BCR reactor by CFD method, different parameters of the fluid are acquired as the CFD outputs parameters. The output parameters consist of the x, y, and z coordinates which denote pressure, air superficial velocity, and air volume fraction, simultaneously. In this study, the CFD outputs were assessed by combining the intelligence optimization algorithm of ant colony and fuzzy inference system (FIS) with.

To use the Ant colony algorithm, part of the CFD outputs were considered as input and the others were considered as output. In this research, five inputs were utilized; the first input was the x coordinate, the second input was y coordinate and the third was the z coordinate. The pressure which was one of the traits of the fluid inside the BCR is the fourth input; air superficial velocity another characteristic of the fluid inside BCR is the fifth input, whereas air volume fraction is considered as output. To initiate the learning process by artificial intelligence (ant colony algorithm), the following conditions are assumed:

The maximum iteration is 100, the total data number is 1500, the value of p represents a percentage of the data that has been used in the learning processes and is considered as %70. In the training process, %70 of the data was involved and %100 of the data was evaluated in the training process. The clustering type was assumed as Fuzzy c-means (FCM). With the above mentioned assumptions, by considering the input of the x coordinates and the output of the air volume fraction, the training and testing processes were performed separately for 20, 30, and 40 numbers of ants. As presented in Fig. 3, the best Regression (R) value is 0.30 for a number of 30 ants which shows that FIS does not have sufficient intelligence in the learning process using the ant colony algorithm, and the change in the number of ants has made no significant enhancement in the FIS intelligence.

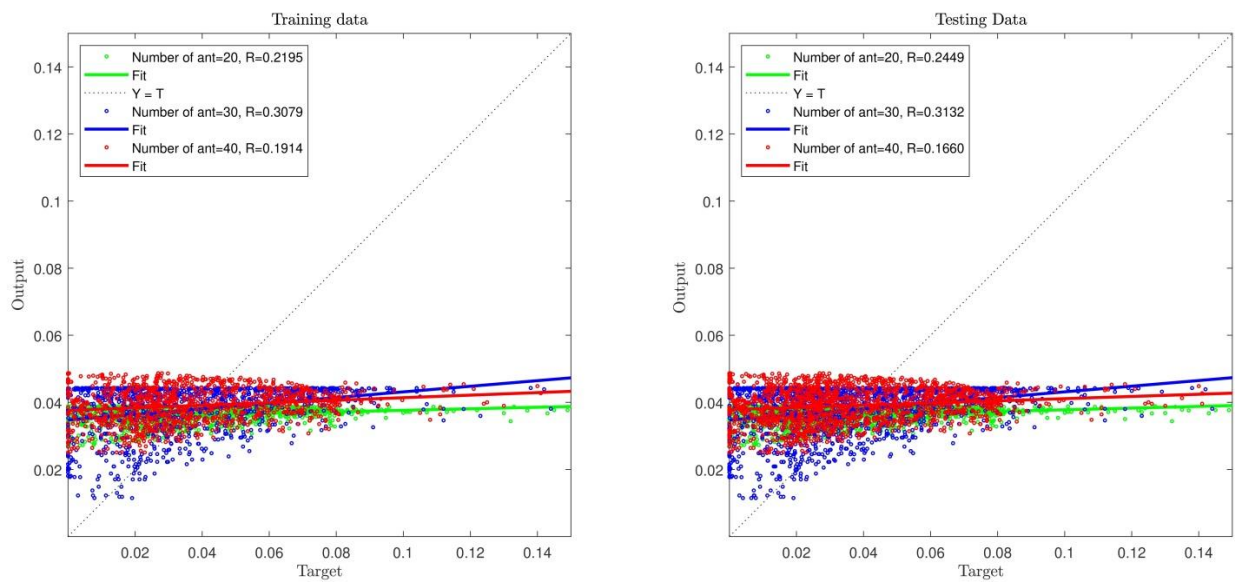

Figure 3: ant colony algorithm training and testing process with one input (number of ant =20, 30, 40; number of data=1500; max iteration=100; P=%70; FCM clustering).

To boost the system intelligence, the number of inputs was increased and evaluated; the x coordinate and y coordinate were considered as inputs, and learning processes were carried out for 20, 30, and 40 ants separately. Fig. 4 doesn't show much enhancement in system intelligence.

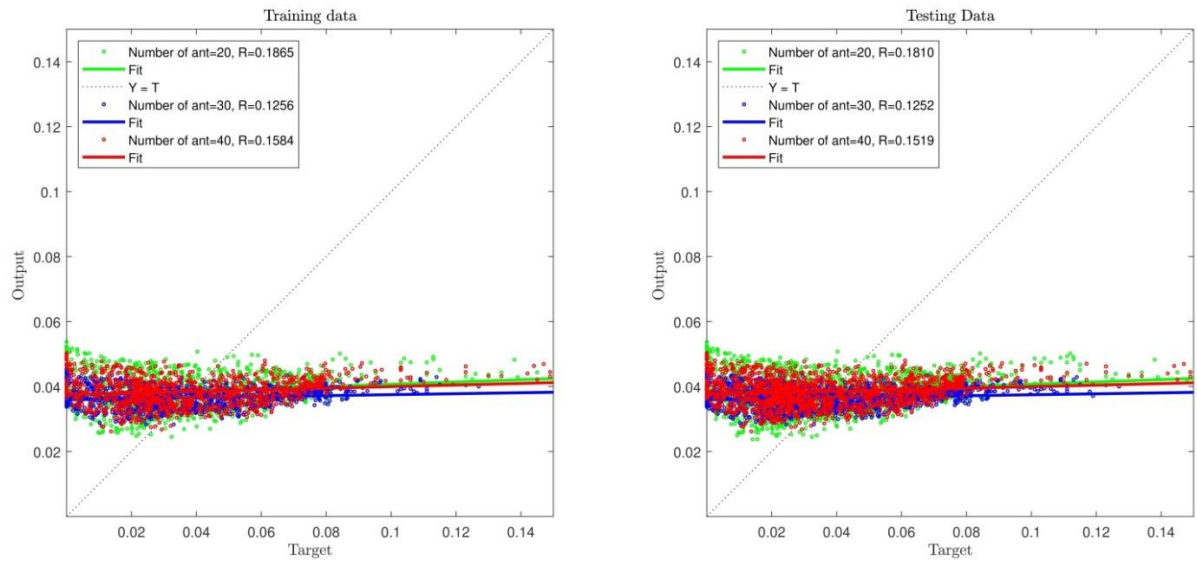

Figure 4: ant colony algorithm training and testing process with two inputs (number of ant =20, 30, 40; number of data=1500; max iteration=100; P=%70; FCM clustering).

To elevate the ant colony algorithm intelligence, the increase in the number of inputs from 2 to 3 was considered and z coordinate was considered as third input and the air volume fractions were considered as output. By conducting separate training and testing procedures for various numbers of ants, no significant changes are observed in intelligence as shown in Fig. 5.

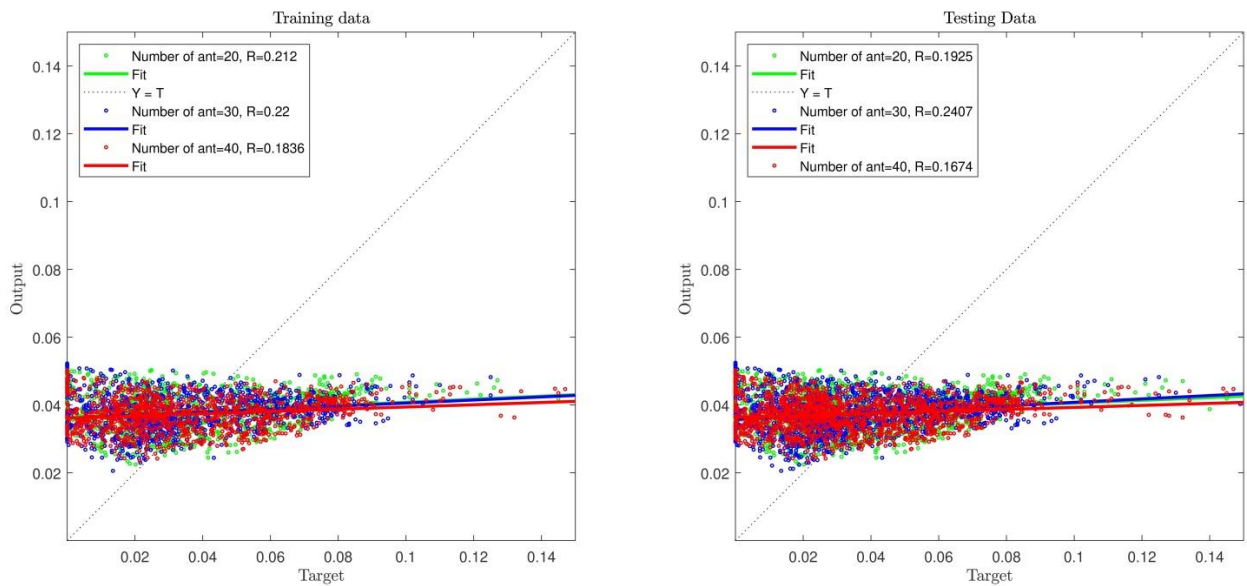

Figure 5: ant colony algorithm training and testing process with three inputs (number of ant =20, 30, 40; number of data=1500; max iteration=100; P=%70; FCM clustering).

In this stage of the study, one of the characteristics of the fluid inside the BCR i.e. pressure was considered as the fourth input. The learning processes (training and testing) for 20, 30, and 40 ants were done separately, but unfortunately, there was still no significant effect on elevating the system intelligence. (See fig. 6)

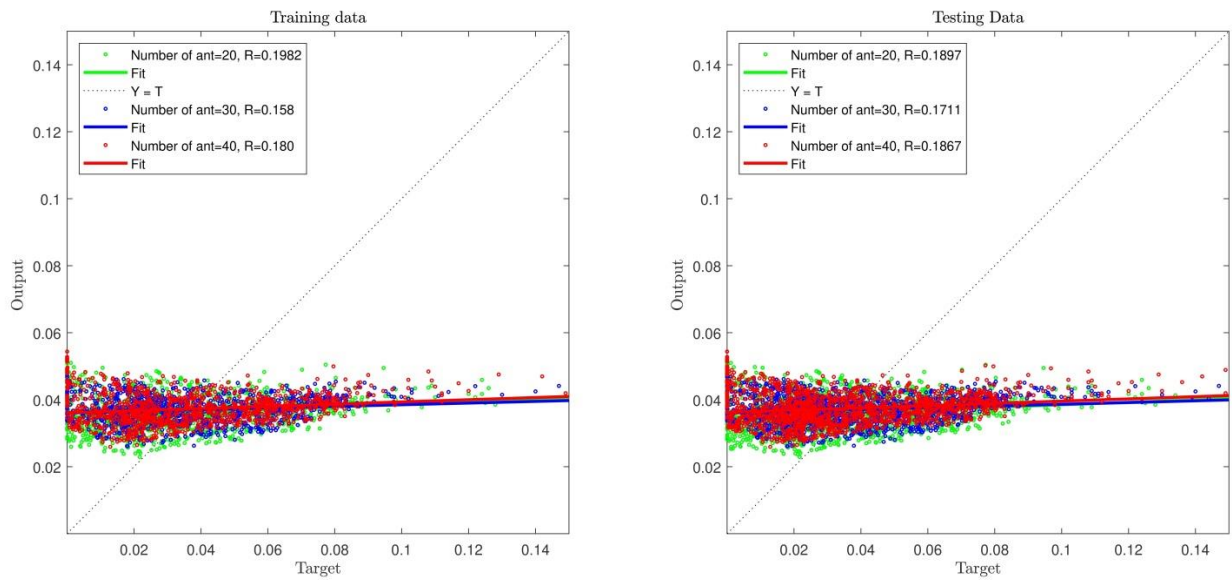

Figure 6: ant colony algorithm training and testing process with four inputs (number of ant =20, 30, 40; number of data=1500; max iteration=100; P=%70; FCM clustering).

Afterward, in order to attain favorable system intelligence, another characteristic of the fluid inside BCR i.e. air superficial velocity was considered as the fifth input, and the learning processes were carried out for 20 ants. As presented in fig.7, the value of R for the training process has increased from about 0.20 to 0.96 and for the testing process, it has increased to 0.95, which indicates a very favorable enhancement in the system intelligence and the achievement of complete intelligence for the system. Using this intelligence, various parts of the BCR can also be predicted. In Fig. 8, points of BCR that participated in the learning process are observed that used in the ant colony algorithm learning process.

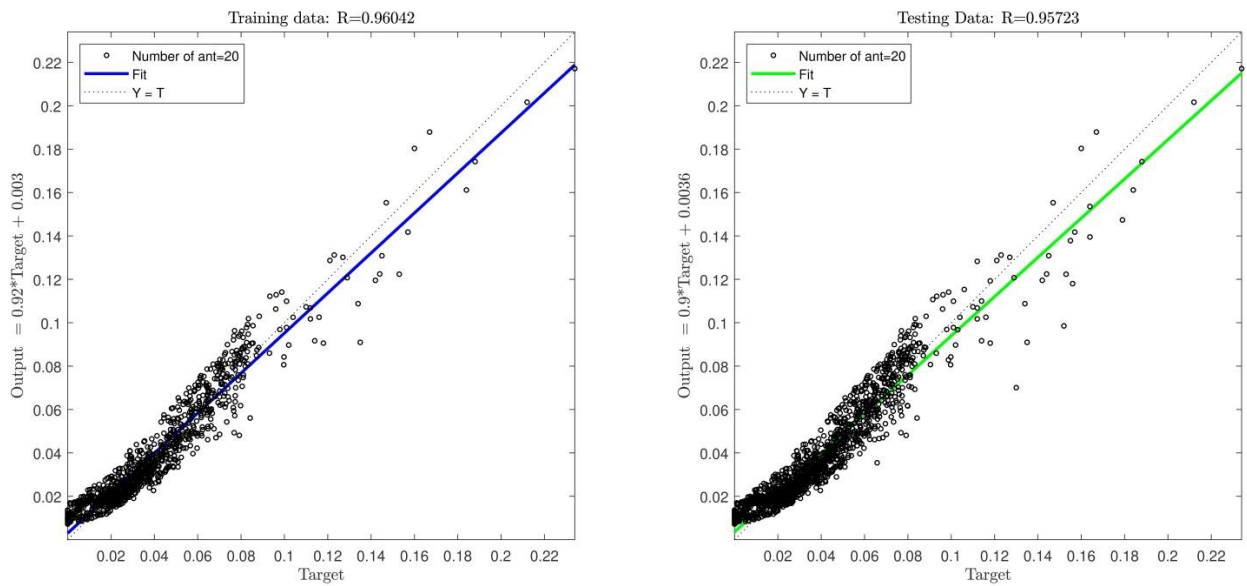

Figure 7: ant colony algorithm training and testing process with five inputs when (number of ant =20; the number of data=1500; max iteration=100; P=%70; FCM clustering).

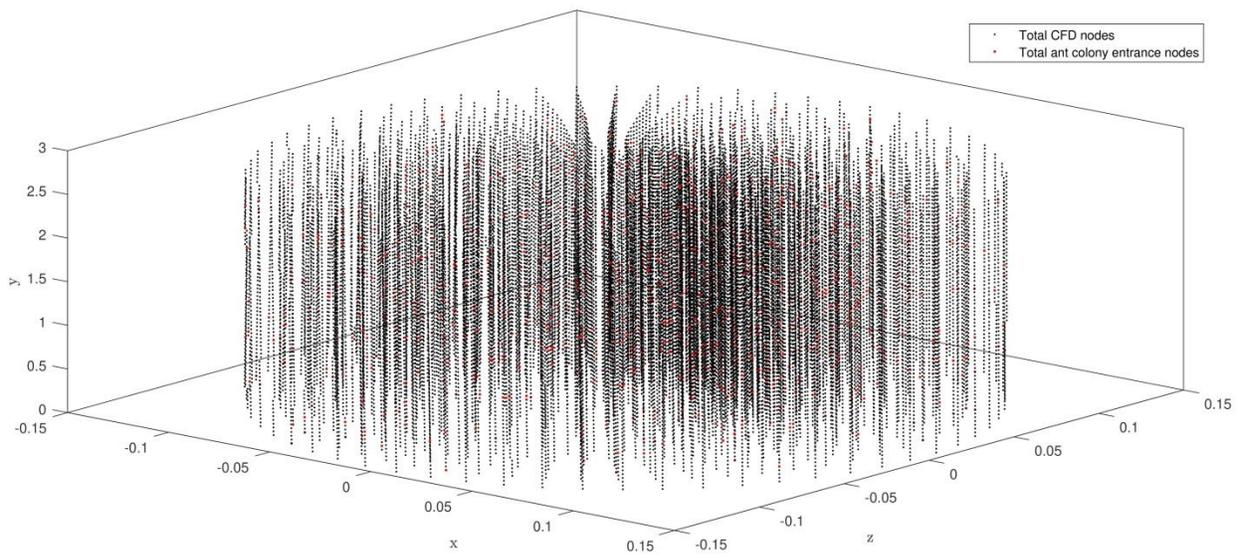

Figure 8: CFD method nodes used in the ant colony algorithm learning process.

The combination of artificial intelligence (ant colony algorithm) and the CFD method decreases the required time for calculations by the CFD method, it also leads to avoiding the solving of complex equations by the CFD method; moreover, by exploiting the created intelligence, much more information and result points can be acquired.

A comparison of the CFD output nodes and ACO algorithm prediction nodes demonstrates a very favorable agreement between the CFD results and the ant colony algorithm output (see Fig. 9(a, b)). Using this obtained intelligence, nodes that are not present in the learning process can be predicted and this shows the very favorable capacity of the artificial intelligence (ant colony algorithm), which is very advantageous and effective (see fig. 10).

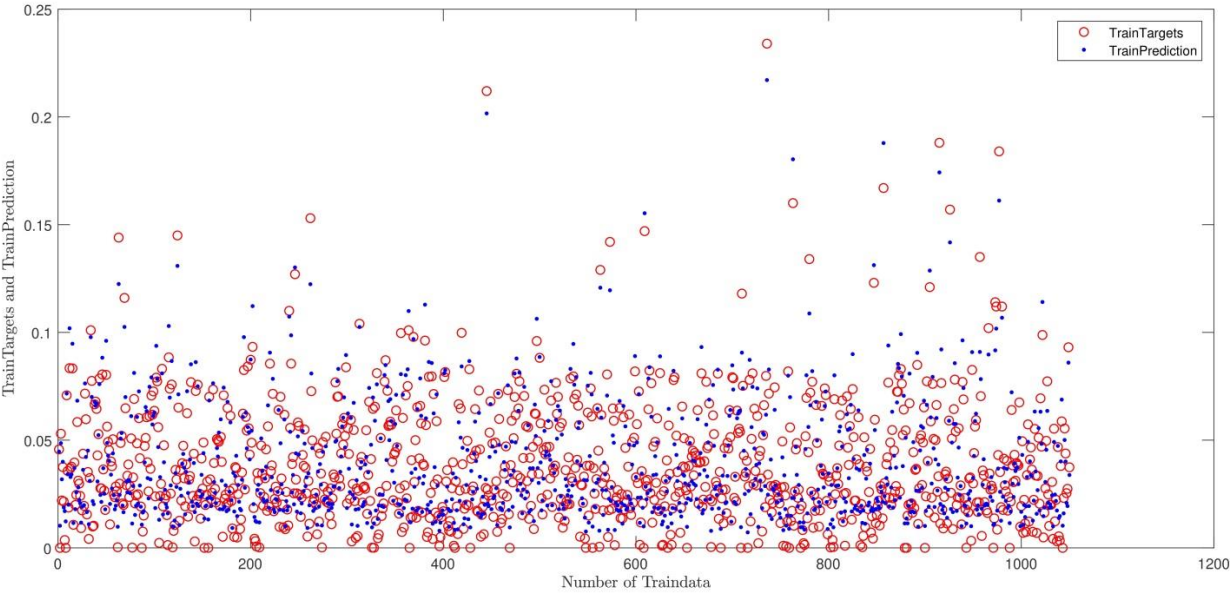

Figure 9(a): Training process target and prediction (number of ant =20; number of data=1500; max iteration=100; P=%70; FCM clustering).

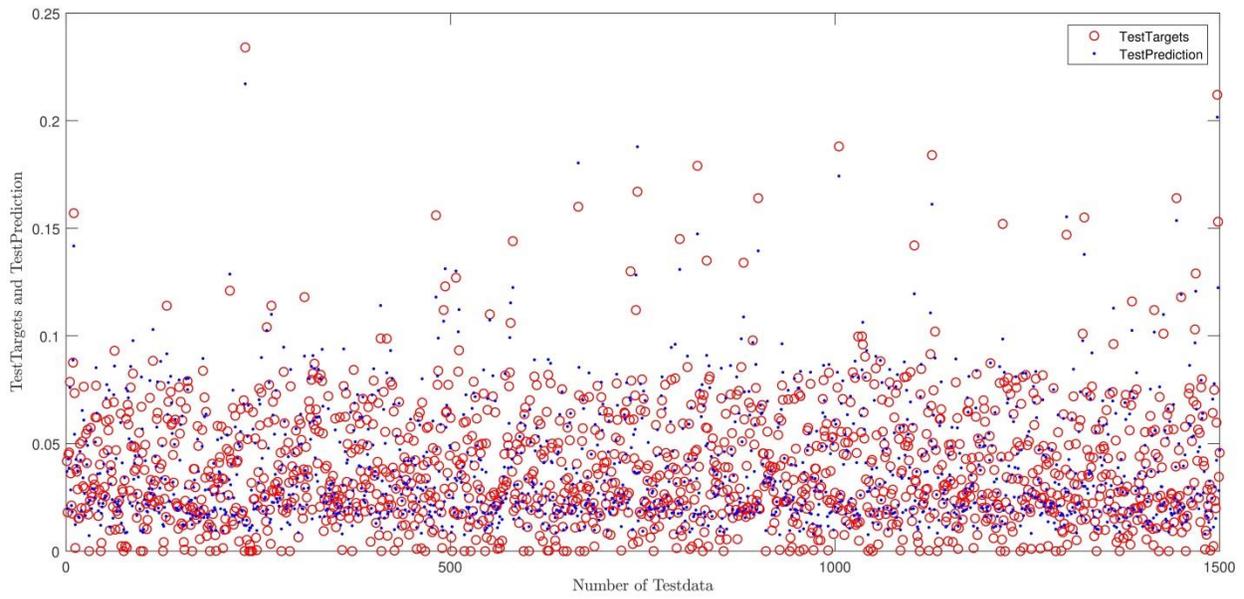

Figure 9(b): Testing process target and prediction (number of ant =20; number of data=1500; max iteration=100; P=%70; FCM clustering).

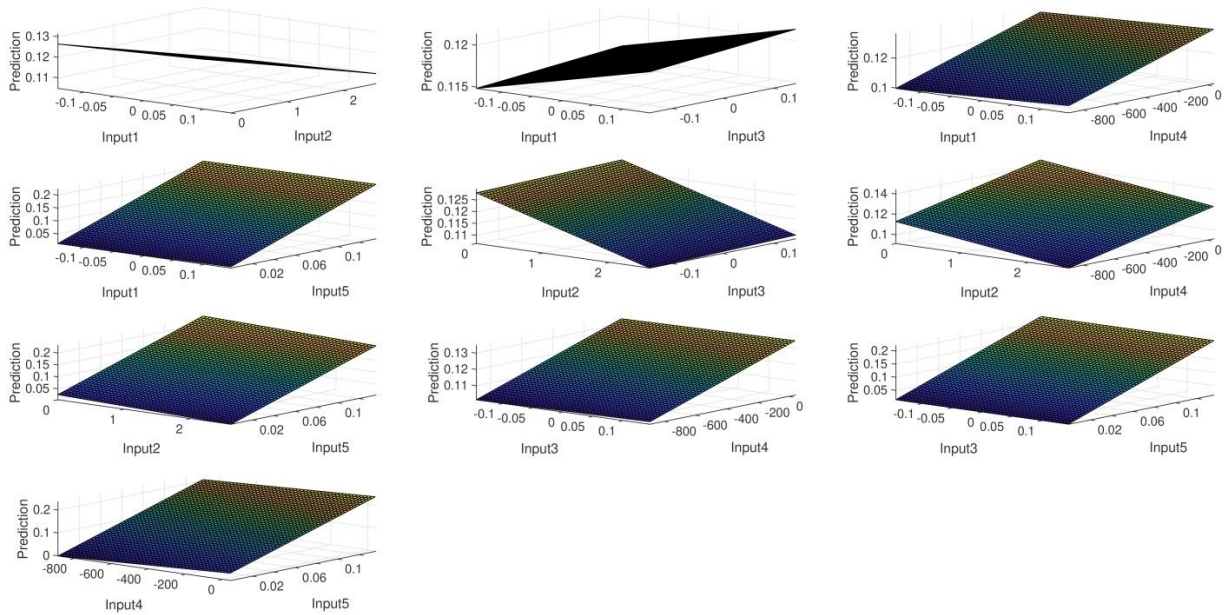

Figure 10: Ant colony algorithm Prediction (number of ants =20; number of data=1500; max iteration=100; P=%70; FCM clustering).

## 4-Conclusion

Current work describes the simulation of the gas fraction based on different bubble column characteristics with ant colony approach. In particular, the CFD data are considered as training inputs of the ant colony method, and this method predicts the behavior of the bubble column reactor. The simulation of the gas fraction is implemented in a 3D domain of fluid-structure and it is compared with the results of CFD. In the training process, the reactor's top bottom and middle levels are chosen for computing the BCR hydrodynamics because of the gas holdup behavior at the mentioned levels. The Ant colony method model is an appropriate tool for prediction with almost 30 percent of data in the learning state. Nevertheless, the tuning parameters of this model significantly enhance the ant colony method's intelligence. Also, it is possible to train it in a highly short period of time (iteration), which provides a quick learning procedure having very small computational time and efforts. Moreover, as no obstacle of computational time is present, a higher amount of data can be generated in the input domain of data indicating novel reactor conditions with no experimental or numerical outcomes. This new perception of data analysis with artificial ants and local search algorithms is sophisticated process for post-processing the data as other researchers started with other soft-computing methods. Prediction of the fluid flow around bubbles can be very complicated and estimation of the vortex structure near bubbles requires more training data. This new combination of CFD and AI can provide more tuning parameters in AI prediction of the reactor to achieve accurate prediction results, and enables us to organize data during training and optimization base on the biological overview. For future studies, other biological optimization methods can be combined with inference fuzzy system to predict the BCR hydrodynamics. However, the ant colony method can be modified based on different ants such as Tapinoma nigerrimum, Redwood ant and Myrmecia.

ACKNOWLEDGMENT

We acknowledge the financial support of this work by the Hungarian State and the European Union under the EFOP-3.6.1-16-2016-00010 project.